\documentclass[conference]{IEEEtran}
\usepackage{caption}
\usepackage{color}

\usepackage{url}
\usepackage{cite}
\usepackage{amsmath,amssymb,amsfonts}
\usepackage{algorithmic}
\usepackage{graphicx}
\usepackage{textcomp}
\usepackage{xcolor}

\def\BibTeX{{\rm B\kern-.05em{\sc i\kern-.025em b}\kern-.08em
    T\kern-.1667em\lower.7ex\hbox{E}\kern-.125emX}}

\begin{document}
%\pagenumbering{gobble}
\title{\textbf{\Large Reliable Fleet Analytics for Edge IoT Solutions}}% \\[-1.5ex] and subtitle}\\[0.2ex]}

% \author{\IEEEauthorblockN{~\\[-0.4ex]\large Emmanuel Raj\\[0.3ex]\normalsize}
% \IEEEauthorblockA{TietoEVRY, AI Center of Excellence\\
% %Georgia Institute of Technology\\
% Keilalahdentie 2-4, 02150 Espoo, Finland\\
% Email: {\tt emmanuelraj7@gmail.com}}
% %\and
% %\IEEEauthorblockN{~\\[-0.4ex]\large Homer Simpson\\[0.3ex]\normalsize}
% %\IEEEauthorblockA{Twentieth Century Fox\\
% %Springfield, USA\\
% %Email: {\tt homer@thesimpsons.com}}
% \and
% \IEEEauthorblockN{~\\[-3.0ex]\large Magnus Westerlund\\ and Leonardo Espinosa-Leal\\[0.2ex]\normalsize}
% \IEEEauthorblockA{    %Department of Business and Analytics\\
% Arcada University of Applied Sciences\\
% Jan-Magnus Janssonin aukio 1, 00560 Helsinki, Finland\\}}

\author{\IEEEauthorblockN{Emmanuel Raj}
\IEEEauthorblockA{Machine Learning Engineer\\
TietoEvry\\
Helsinki, Finland\\
emmanuelraj7@protonmail.com\\}
\and
\IEEEauthorblockN{Magnus Westerlund}
\IEEEauthorblockA{
Arcada University of Applied Sciences\\
Helsinki, Finland \\
magnus.westerlund@arcada.fi\\}
\and
\IEEEauthorblockN{Leonardo Espinosa-Leal}
\IEEEauthorblockA{
Arcada University of Applied Sciences\\
Helsinki, Finland \\
leonardo.espinosaleal@arcada.fi\\}
}

\maketitle

\begin{abstract}
%\boldmath
In recent years we have witnessed a boom in Internet of Things (IoT) device deployments, which has resulted in big data and demand for low-latency communication. This shift in the demand for infrastructure is also enabling real-time decision making using artificial intelligence for IoT applications. Artificial Intelligence of Things (AIoT) is the combination of Artificial Intelligence (AI) technologies and the IoT infrastructure to provide robust and efficient operations and decision making. Edge computing is emerging to enable AIoT applications. Edge computing enables generating insights and making decisions at or near the data source, reducing the amount of data sent to the cloud or a central repository. In this paper, we propose a framework for facilitating machine learning at the edge for AIoT applications, to enable continuous delivery, deployment, and monitoring of machine learning models at the edge (Edge MLOps). The contribution is an architecture that includes services, tools, and methods for delivering fleet analytics at scale. We present a preliminary validation of the framework by performing experiments with IoT devices on a university campus's rooms. For the machine learning experiments, we forecast multivariate time series for predicting air quality in the respective rooms by using the models deployed in respective edge devices. By these experiments, we validate the proposed fleet analytics framework for efficiency and robustness.
\end{abstract}

\begin{IEEEkeywords}
Fleet Analytics; Edge Computing; Machine Learning; Internet of Things; AI
\end{IEEEkeywords}

\IEEEpeerreviewmaketitle

\section{Introduction}

In the last years, we have seen a surge in cloud computing, making it a vital part of businesses and IT infrastructures. The paradigm offers benefits to organizations such as no need to buy and maintain infrastructure, less technical in-house expertise required, scaling,  robust services, and pay as you go features. Organizations can now centrally store massive amounts of data and optimize computational resources to deliver on their data processing needs, which depict the change from localized computing (own servers and data centers) to centralized computing (in the cloud). Cloud computing is today an industry that has enabled many new opportunities in terms of computation, visualization, and storage capacities \cite{kratzke2017understanding}. However, cloud computing has also introduced significant security and data privacy issues and challenges \cite{singh2016survey}; it is essential to critically assess limitations, alternative designs, and develop an overall understanding of ecosystem design \cite{popovic2010cloud}.

With the advent of big data, mobile devices (self-driving cars, mobiles, etc.), and industrial IoT, there is now an increasing emphasis on local processing of information to enable instantaneous decision making. We are witnessing a shift in trend from conceptually centralized cloud computing to decentralized computing. Here, \emph{Edge Computing} is the process of performing computing tasks physically close to target devices, rather than in the cloud \cite{shi2016edge,10.1145/3316782.3321546}. It enables extracting knowledge, insights, and making decisions near the data origin quickly, secure, and local, which facilitates decentralized processing. Edge computing also enables data confidentiality and privacy preservation, something that is becoming essential across multiple industries. The growing amount of (IoT) data and the associated limitations of using cloud computing (networking, computation, and storage) are currently drivers for decentralized systems, such as Edge Computing.

To achieve a computing approach that considers resource optimization in terms of energy, efficiency, operational costs, and human resources, we need a shift from pure cloud computing to a more nuanced architecture that provides sustainable computing resources and infrastructure for organizations to run their services \cite{bilal2018potentials,blogEmmanuel8enablers}. Green IT, where energy and resource optimization are essential, has also been extended to Green IoT \cite{ shaikh2015enabling}. Hence, we see investments from the public and private sectors going towards building smart solutions and cities that enable smart societies \cite{bilal2018potentials}. In use-cases where sensitive data is handled or require low latency delays, cloud computing may not be a perfect solution. 

With examples such as big data, self-driving cars, and IoT, there is an increasing emphasis on local processing of information to enable instantaneous decision making using AI, also called the Artificial Intelligence of Things (AIoT) \cite{iot-future,aiot}. Edge computing can unlock the potential for making real-time decisions or extract knowledge near the data origin in a resource-efficient and secure manner  \cite{shi2016edge}. Edge computing has gradually emerged from the client/server architecture; for example, in the late 1990s \cite{genesis-akamai} showed how resource constrained mobile devices could offload some of their processing needs to servers. Later the Content Delivery Network (CDN) was launched by Akamai \cite{bosch} and certain notorious peer-to-peer networks. Since then, there have been major developments in cloud computing, edge computing, IoT, and low latency networks. When Akamai launched its CDN the idea was to introduce nodes at locations geographically closer to the end-user to deliver cached content such as images and videos. Today many companies utilize a similar approach for speech recognition services and other AI-enabled or processing heavy services. 
 
A massive growth in IoT device deployments, as of 2018, there was an estimated 22B devices \cite{ericsson-new-prediction}, has not happened without significant security challenges. To manage the scale of IoT device deployments, edge computing will play an important role. The aim is to promote IoT scalability and robustness in order to handle a huge number of IoT devices and big data volumes for real-time low-latency applications while avoiding introducing new security threats. Edge computing is increasingly defined as performing data processing on nearby compute devices that interface with sensors or other data origins \cite{shi2016edge}. Edge-based IoT solutions must cover a broad scope of requirements while focusing on scalability and robustness through resource distribution. 

The structure of the paper is the following. Section \ref{sec:scalability}, expounds the design demands for creating Artificial  Intelligence of  Things. In Section \ref{sec:operational}, we review the AIoT design support methodologies and practices. Section \ref{sec:fleet}, defines our modular design framework for fleet analytics, and in Section \ref{sec:experimental}, we discuss a validation of our framework. Section \ref{sec:conclusion} concludes the paper with a note about future work.

 \begin{figure}[tp]
    \centering
    \captionsetup{justification=centering}
    \includegraphics[width=9cm]{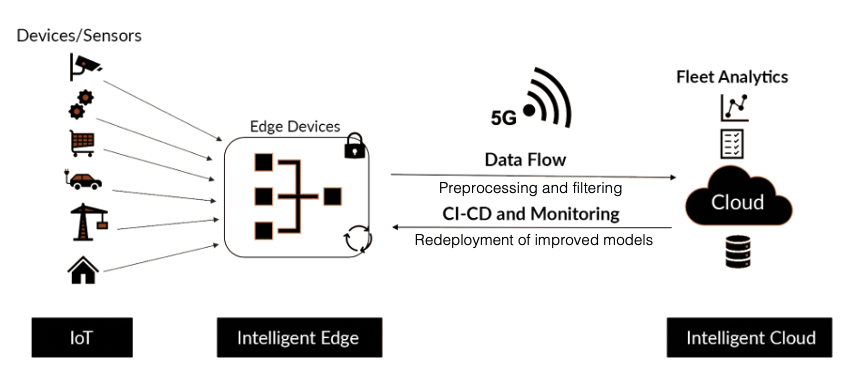}
    \caption{Intelligent edge and Intelligent cloud powered by 5G networks}
    \label{Figure:high level architecture}
\end{figure}

\section{Scalability and Reliability for AIoT}\label{sec:scalability}

 In order to perform computing close to the data source and to offload centralized computing to a decentralized infrastructure, require explicit and well formalized processes. Edge computing means we should apply different machine learning algorithms at the edge, enabling new kinds of experiences and new kinds of opportunities across many industries, ranging from mobility, connected home, security, surveillance, and automotive. Further, edge computing may also enable secure and reliable performance for data processing and coordination of multiple devices \cite{beyer2018site}. Figure \ref{Figure:high level architecture} depicts an overview diagram of how a secure and reliable intelligent edge architecture is constructed.

Reliability for distributed systems demands strict protocols that each node adheres to. Reliability, as defined by Adkins \emph{et al.} \cite{googleSecRel}, is considered a distinct topic from security, although sharing several properties. Reliability is a demanding task that must be considered early in the planning phase to capture the emerging properties and continuously capture requirements for achieving reliability that may evolve in time. Reliability for today's landscape involves other considerations than purely technical ones. The main driver for reliable edge solutions may be the increase of regional legislation in the digital space \cite{cloudflare}. 

IoT systems' distributed nature means that dependencies between nodes should be avoided while striving for integrating automated redundancy when designing systems. In Table~\ref{designConsiderations}, we summarize some of the considerations for building edge IoT solutions that include fleet analytics. Fleet analytics is still an emerging field of research, and in the absence of direct references, we provide general references for each topic.

\captionsetup{font={footnotesize,sc},justification=centering,labelsep=period}%
\begin{table}[tbp]
\caption{Design reasons and considerations for utilizing fleet analytics for edge IoT solutions.}\label{designConsiderations}
\centering%
\begin{tabular}{p{1.3cm}p{5.4cm}p{0.8cm}}
\hline
\textit{Concept} & \textit{Description} & \textit{Reference} \\
\hline

Local compliance & Regional regulations e.g. for privacy and security may be easier to implement with localized computing. & \cite{cloudflare}\\

Service level & Meeting service level objectives for IoT networks may require precise measurements at the edge to monitor decision making and feedback loops on the physical plane. & \cite{beyer2018site}\\

Ease of use & Building a reliable decoupled system may require a design where data is processed close to the IoT node. Thus, avoiding transferring data to a different backend environment. & \cite{cloudflare}\\

System stability & Stability under heavy load demands scalability and throughput, for distributed systems this means that single point of failure designs must be avoided. & \cite{shi2016edge}\\

System safety & Systems that interact with their surroundings may benefit from physical proximity to models and supervising algorithms in order to speed up decision making. This demands well-formed streaming pipelines that consider freshness, correctness, and coverage. & \cite{beyer2018site}\\
\hline
\end{tabular}
\end{table}
\captionsetup{font={footnotesize,rm},justification=centering,labelsep=period}%

In Table~\ref{planes}, we separate the design considerations further into three different planes. First, the hardware plane that the IoT device is implemented on. Here we should note that a multitude of designs exist, some with considerable processing power limited mainly by a thermal dissipation to systems on a chip (SOC) running on battery power. The second plane is represented by the AI models processing the data and interactions that the IoT device captures. These models are susceptible to drift among many other issues, meaning that both the input and output should always be monitored for any statistical abnormalities. Third, is the service plane where decision making and reliability automation come together. 

\captionsetup{font={footnotesize,sc},justification=centering,labelsep=period}%
\begin{table}[tp]
\caption{Design planes for fleet analytics in edge IoT solutions.}
\label{planes}
\centering%
\begin{tabular}{p{1.7cm}p{6.3cm}}
\hline
\textit{Plane} & \textit{Description} \\
\hline
Hardware & Telemetry from devices and their sensors may help us monitor the device itself and the environment the device resides in.\\

AI & The use of machine learning means that the systems must be continuously monitored during their operation.\\

Service & Operational support methods help to deploy and maintain a reliable fleet analytics solution.\\
\hline
\end{tabular}
\end{table}
\captionsetup{font={footnotesize,rm},justification=centering,labelsep=period}%

\section{Operational support methodologies}\label{sec:operational}

To understand the need for Fleet analytics is vital to turn an eye to software development practices starting from DevOps to DataOps to MLOps. 

\subsection{DevOps}

DevOps extends Agile development practices by streamlining software changes through the build, test, deploy, and delivery stages. DevOps empowers cross-functional teams with the autonomy to execute on their software applications, driven by continuous integration, continuous deployment, and continuous delivery. It encourages collaboration, integration, and automation among software developers and IT operators to improve efficiency, speed, and quality of delivering customer centric software. DevOps provides a streamlined software development framework for designing, testing, deploying, and monitoring production systems. DevOps has made it possible to ship software to production in minutes and keep it running reliably \cite{bass2015devops}.

\subsection{DataOps}

DataOps refers to practices centered around data operations that bring speed, agility, and reproducibility for end-to-end data pipelines. The DataOps process considers the entire data life cycle activities and is derived from DevOps. The business aim of DataOps is to achieve data quality from optimized data pipelines by utilizing automated orchestration and monitoring of processes. DataOps practices assume that data will be processed further in various analytics-based setups~\cite{rajad2020}.

\subsection{MLOps}

\begin{figure}[t]
\centering%
\includegraphics[width=8.7cm]{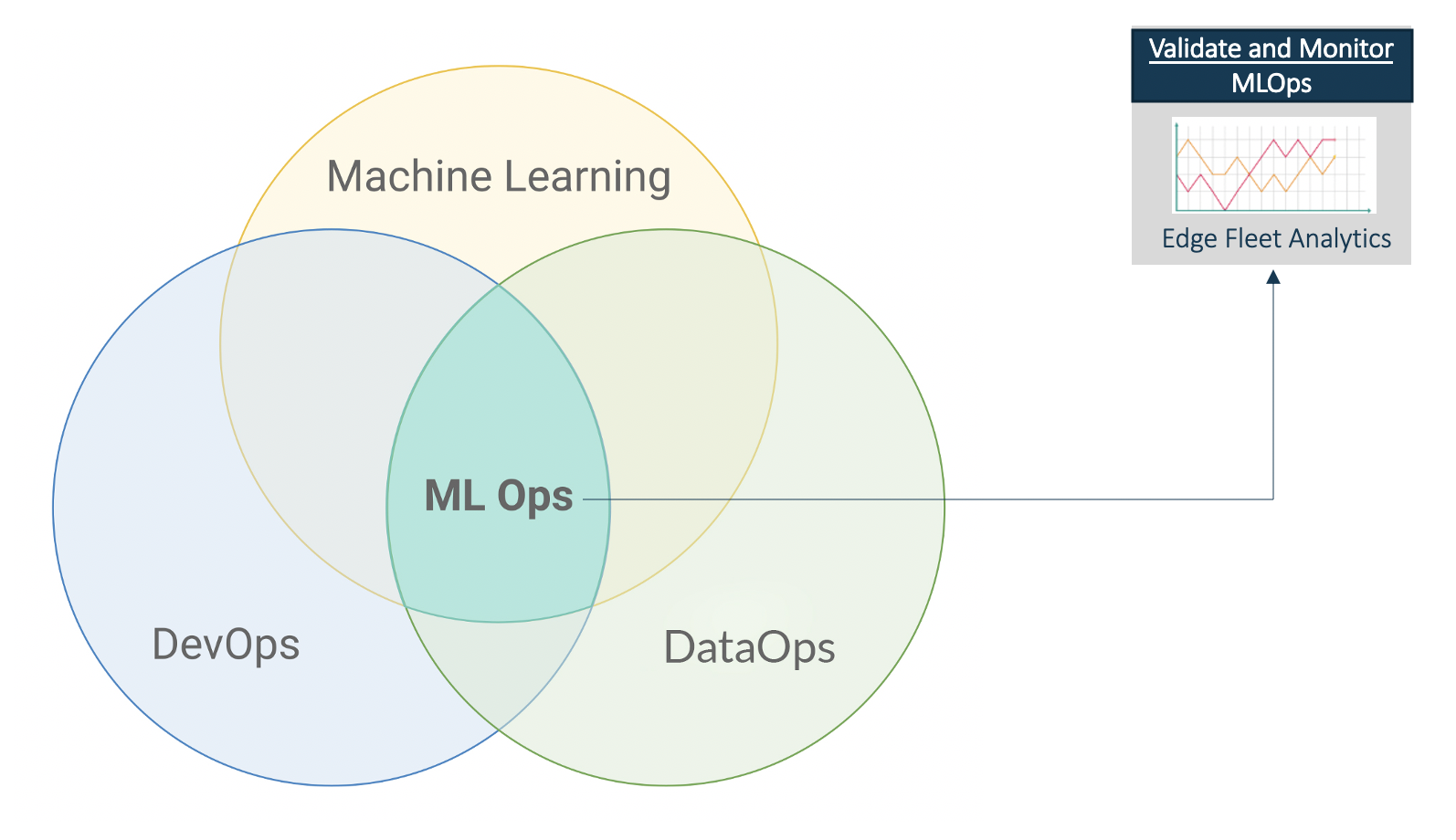}
\caption{Need for Edge Fleet Analytics Framework}
\label{Figure:need}
\end{figure}

Software development is an interdisciplinary field and is evolving to facilitate machine learning in production use. MLOps is an emerging method to fuse machine learning engineering with software development. MLOps combines Machine Learning, DevOps, and Data Engineering, and aims to build, deploy, and maintain machine learning models in production reliably and efficiently. Thus, MLOps can be expounded by this intersection, as depicted in Figure \ref{Figure:need}. MLOps was defined in \cite{banerjee2020challenges} as 1) dealing with continuous training and serving, 2) monitoring solutions, 3) high level of automation, and 4) an orchestrated environment for model validation. MLOps is still only an emerging operational support method. However, the need to establish operational trust towards ML models and integrate machine learning with software development speaks in MLOps favor.   

\section{Fleet Analytics for IoT Networked Devices}\label{sec:fleet}

To manage distributed IoT systems (aka. fleet management), we have implemented a fleet analytics framework that allows us to address the three different operational support methodologies in a unified way. Fleet analytics for distributed IoT systems arises from the necessity to continuously validate and monitor the operational methods whose distributed nature makes them somewhat different from traditional development. Thus, we introduce a robust and reliable fleet analytics framework that can be used in production environments. 

\begin{figure}[htb]%p]
\centering%
\includegraphics[width=8.7cm]{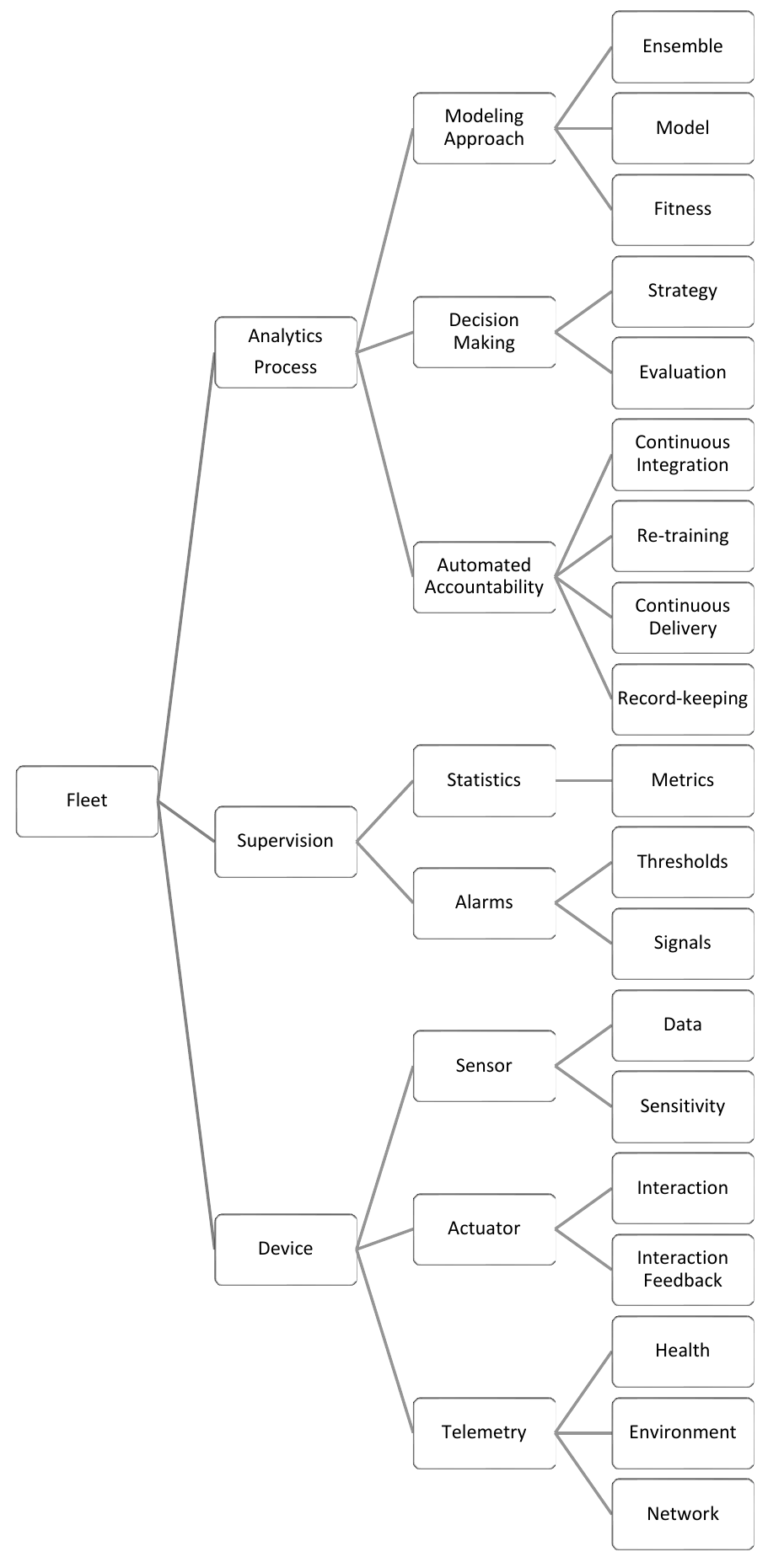}
\caption{A modular design framework for a fleet analytics system.}
\label{taxonomy}
\end{figure}

Fleet analytics enables validation and monitoring of edge devices (via telemetry data), sensor data, and machine learning models. Fleet analytics provides a continuous holistic and analytical view of the health of the system. The aim has been to automate the monitoring and orchestration of devices. An important goal has been to create a framework for fleet analytics that maintains high reliability for the system. In Figure~\ref{taxonomy}, we propose a modular design framework. We want to acknowledge that the framework is still a work in progress and is not complete. The proposed framework intends to clarify the design components of the proposed system.

\subsection{Framework proposal}

The framework proposes a triune approach to fleet analytics for edge computing driven by MLOps. To validate and monitor the edge computing system is vital to monitor the analytics process, supervision (system actions and performance), and device health.

\subsubsection{Analytics Process}
The analytics process is key to driving the decisions and actions of the system. Hence it is vital to monitor the analytics process end-to-end. This means starting from data processing, training the machine learning model, deploying and monitoring the models on edge devices. We have separated the analytics process into three operations: the \emph{modeling approach}, the \emph{decision making}, and the continued upkeep that we refer to as \emph{automated accountability}. To synchronize these three operations, MLOps provides a method for orchestrating the transfer of machine learning models in the system and to devices, while also assisting in the continued monitoring of the system. MLOps empowers data scientists and application developers to develop and bring machine learning models to production, that for an edge setup like ours, means that models may be trained on shared, dedicated machinery. At the same time, the inference is performed at the outermost edge close to the recording sensor or actuator. MLOps thereby enables a systematic approach to track, version control, audit, certify, and re-use every asset in the ML life cycle. By providing orchestration services for infrastructure, MLOps streamlines the life cycle management of edge solutions. To track and monitor the analytics process as part of fleet analytics holistically, we observe these following aspects:

\paragraph{Modeling approach} This aspect of the analytics process defines the machine learning model setup and enables training, evaluation, and testing (fitness) for production. In some instances, it may involve ensembles and arranging models logically to specify well formed processing pipelines. 

\paragraph{Decision making} The modeling approach utilizes a set of query inputs and produces inference from experience stored in the knowledge base or training data (used to train the models). The decision making operations enables the system to interact with the environment and to introduce expert knowledge. For high-impact decisions, such as automated system operation, that can impact human well-being or damage property or the environment, it is prudent to introduce fail-safe measures so that the model output is confined within a trusted decision space. The key to good decision making is defining a decision making strategy that includes planning, formulation, implementation of various methods, and workflows. When a decision making strategy is implemented, it is essential to track and monitor the progression over time to ensure an efficient and reliable performance for the complete system.  

\paragraph{Automated Accountability} When the human element is introduced into the design of decision support systems, entirely new layers of social and ethical issues emerge but are not always recognized as such. Hence, automating operations is intended to reduce these issues and the dependence on human ad-hoc interaction. Some key drivers of automation are continuous integration and continuous deployment because they enable the ability to automate model retraining and deployment of the latest models according to the latest system developments and data. Such practices should reduce the occurrence of human error or need to maintain direct human oversight of system developers. With proper auditing and record-keeping, it is efficient to monitor and debug the system's continued operations. 

\subsubsection{Supervision} 
Having a reliable supervision strategy in place is vital for the efficient functioning of machine learning driven systems. Systems are supervised statistically using metrics defined to monitor the performance. As decision making is an essential behavior of an analytics-based system, decisions also need to be supervised and monitored to avoid any unnecessary failures and harmful system interactions. System alarms can be created for critical decisions or failures using thresholds and signals. Such alarms can provide human supervisors with an asynchronous method for ensuring robust system performance.  

\subsubsection{Device}
There are typically several types of devices in a complete system; here we reduce the types to three different types. Sensors that provide measurement data of the environment, actuators that perform actions, and telemetry data sources that can measure both physical and virtual properties that provide meta information about the functioning system. DataOps practices can be used to automate data collection and provide reproducibility and end-to-end data pipelines.

Monitoring the health and performance of edge and IoT nodes is essential to avoid any system's unexpected failures. Telemetry data from the nodes is an important part of fleet analytics. Telemetry data ensures that the devices are running as intended and that any potential failures can be predicted in advance and addressed before they occur. Telemetry data offers diagnostic insights into the device health, environment, and network. This data provides valuable insight into the health and environment of the IoT devices, actuators, and edge devices, which can be used to automate much of their operation through fleet analytics. As we consider, Fleet analytics is not complete without comprehensive device data in the form of telemetry data for ensuring data quality and integrity. This is also a reason for considering DataOPS as an operational support method for fleet analytics.

\section{Experimental Framework Validation}\label{sec:experimental}
\label{experiments}

\begin{figure}[tbp]
\centering%
\includegraphics[width=8.7cm]{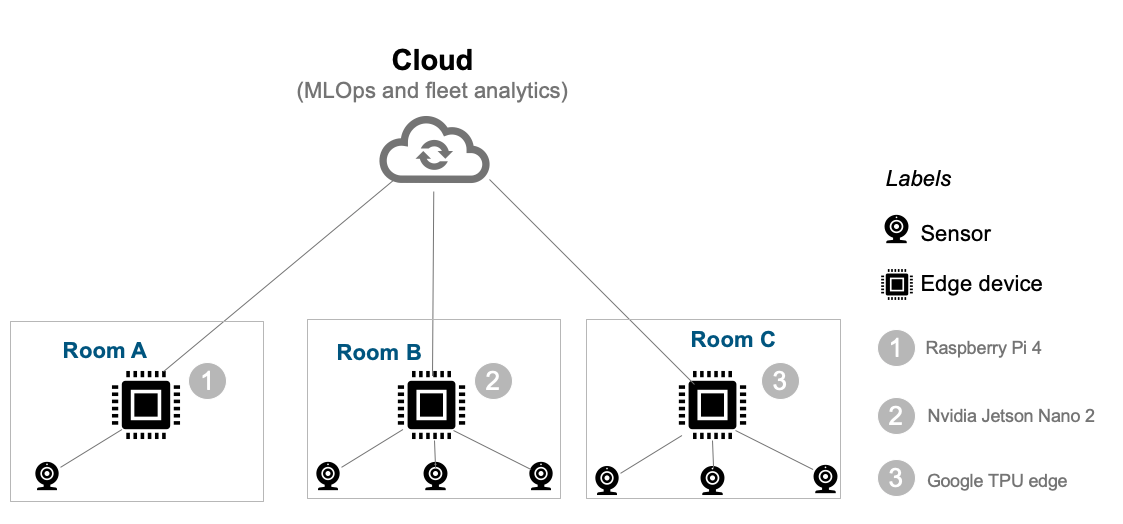}
\caption{Experimental setup}
\label{experiment_setup}
\end{figure}

To validate the fleet analytics framework and design, we have implemented a system and conducted a live experiment for 45 days. We use three IoT devices and three edge devices for performing inference from machine learning models to predict the air quality inside three rooms during this process. Each room had one IoT device or sensor that measured the room’s air quality conditions and one edge device to deploy the ML models to and for predicting the changes in the air quality (see Figure \ref{experiment_setup}).

Machine learning models were trained based on three months of historical data from each room, and posteriorly they are deployed on the edge devices in the rooms. The machine learning models used were Multiple Linear Regression (MLR), Support Vector Regressor (SVR), Extreme Learning Machines (ELM), and Random Forest Regressor (RFR). The goal was to predict air quality 15 minutes into the future, inside each room.

\subsection{Analytics Process}

In this subsection, we discuss in detail the analytics process following our design framework. The experiment included data processing, training of machine learning models, deployment of machine learning models, and the monitoring of models on edge devices.

\subsubsection{Modeling Approach} 
In the experiment, we perform multivariate time-series analyses to predict the air quality 15 minutes into the future inside a particular room. With this information, building maintainers could be alerted of possible lousy air quality that needs to be addressed to provide a positive experience for people in the room. For the time being, there is not an integration of the experimental setup with an actuator or the building HVAC system. The collected raw data was sampled every 5 minutes and assembled from 3 months before the experiment. Data column descriptors are listed below. Table \ref{tab:room descriptive analytics} provides some descriptive measures for the data set.

The data descriptors for data collected from IoT devices and their respective data types are shown below:

\begin{itemize}
\item \emph{timestamp}  - Sampling time (datetime)
\item \emph{name} - Name of sensor (str)
\item \emph{room} - The room where the sensor is placed or origin of the data (str)
\item \emph{room type} - Type of room (str)
\item  \emph{floor}	- Floor where data was generated (str)
\item \emph{air quality}	- Air quality index altered (float)
\item \emph{air quality static} - Air quality index unaltered (float) 
\item \emph{ambient light} -	Light level in the room (float)
\item \emph{humidity} -	Humidity in the room (float)
\item \emph{iaq accuracy} -  Indoor Air Quality index altered (float)
\item \emph{iaq accuracy static}	-  Indoor air quality index unaltered (float)
\item \emph{pressure} -	Pressure in the room (float)
\item \emph{temperature} - Temperature in the room (float)
\end{itemize}

\begin{table}[tbp]
    \caption{DESCRIPTIVE STATISTICS FOR AIR QUALITY INDEX (RANGING FROM 0-500) IN SELECTED ROOMS.}
    \centering
    \begin{tabular}{ |p{1.5cm}||p{2cm}|p{1cm}|p{1cm}|  }
     \hline
     \multicolumn{4}{|c|}{Selected Rooms} \\
     \hline
     Room name & Room type & Unhealthy air quality frequency & Avg. air quality index (AQI)\\
     \hline
     Room A10   & Office room    &2033 &   61.92\\
     Room A29 &   Meeting Room  & 2205   &61.40\\
     Room A30 & Meeting Room & 1085&  55.45\\
     \hline
     
    \end{tabular}
    \label{tab:room descriptive analytics}
\end{table}

%\textcolor{red}{
After assessing each room’s air quality time-series data, no trend or seasonality was observed in air quality data for any room. However, there is a change over time in the mean, variance, and covariance. To proceed, we extract meaningful features by performing feature analysis and selection.

\textbf{Feature Extraction:}
After exploring data and identifying patterns, we found some data parameters or columns that were correlated to the air quality in the rooms. Based on the data analysis, we chose the following parameters or columns for training the machine learning algorithms: \emph{air quality static}, \emph{ambient light}, \emph{humidity iaq accuracy static}, \emph{pressure}, and \emph{temperature}. In order to predict air quality, we added a label column \emph{future air quality}  by shifting the column \emph{air quality static} three rows ahead. We also performed a standardization technique for feature scaling, that re-scales the feature value so that it has a distribution with 0 as the mean value and the variance equals 1. With these new features and scaled data, we were ready to start training our machine learning model.

% \textbf{Feature Correlation}
%  To understand the relationship, we observed data and feature correlation between the variable to predict and other attributes in the data. For the feature "future\_air\_quality"  we calculated feature scores using Pearson correlation. 
 
%  \begin{figure}[htp]
%     \centering
%     \captionsetup{justification=centering}
%     \includegraphics[width=9cm]{images/feature_correlation.png}
%     \caption{Feature correlation using pearson correlation.}
%     \label{Figure:Feature-importance}
% \end{figure} 
 
%  We observed patterns for each room for our experiment as shown in figure \ref{Figure:Feature-importance}. The feature "future\_air\_quality" shows a positive correlation with air\_quality\_static, pressure, ambient\_light and iaq\_accuracy to some extent. Positive correlation implies that feature A increases then feature B also increases or if feature A decreases then feature B also decreases. Both features have a linear relationship and move in tandem. In figure \ref{Figure:Feature-importance} we see a strong positive correlation for feature future\_air\_quality to air\_quality\_static and pressure. Humidity has a strong negative correlation which implies if feature A increases then feature B decreases and vice versa.
 
%  \textbf{Feature Scaling}: 
%  We performed standardization technique for feature scaling, It re-scales a feature value so that it has distribution with 0 mean value and variance equals to 1. With this we are ready for machine learning training with our new features and scaled data.

 \textbf{Model Training:} 
We trained four machine learning models on the historical data to predict a future air quality value 15 minutes into the future. To train the models, we perform a 10-fold cross-validation. After assessing each model's performance models were ranked based on performance and is presented here in ascending order: 

\begin{enumerate}
    \item Multiple Linear Regression (MLR)
    \item Support Vector Regressor (SVR)
    \item Extreme Learning Machines (ELM)
    \item Random Forest Regressor (RFR)
\end{enumerate}

\begin{table}[]
    \caption{MODEL TRAINING RESULTS.}
    \centering
\begin{tabular}{ |p{1.5cm}||p{1cm}|p{1cm}|p{1cm}|  }
 \hline
 \multicolumn{4}{|c|}{Model Training Results} \\
 \hline
 Room name & Algorithm & Cross Validation RMSE (train) & Test RMSE\\
 \hline
 Room A10 &   MLR  & 5.020   &5.875\\
 Room A10   & ELM    &6.325 &   6.208\\
 Room A10   & RFR    &10.710 &   9.987\\
 Room A10   & SVR    &6.046 &   5.977\\
 Room A29 &   MLR  & 5.362   &4.158\\
 Room A29 &   ELM  & 11.202   &4.223\\
 Room A29 &   RFR  & 11.676   &9.208\\
 Room A29 &   SVR  & 8.073   &4.176\\
 Room A30 & MLR & 3.648&  3.551\\
 Room A30 & ELM & 7.920&  3.895\\
 Room A30 & RFR & 9.686&  7.720\\
 Room A30 & SVR & 5.177&  3.55\\
 \hline
\end{tabular}
    \label{tab:model training results}
\end{table}

\textbf{Model packaging:}
To make machine learning inference at the edge and resource-heavy training on dedicated hardware, we have to orchestrate the artifacts by serializing, packaging, and redistributing them to where they are needed. The two primary artifacts considered here are:

\begin{itemize}
    \item We used a standardization technique for feature scaling to transform our training data. Similarly, we have to scale incoming input data for model inference to predict future air quality. For this purpose, we serialized the feature scaling object to a pickle file (.pkl).

    \item Machine learning models: All trained and retrained ML models are serialized in the Open Neural Network Exchange (ONNX) format. ONNX is an open ecosystem for interoperable AI models. This means serialization of ML and deep learning models into a standard format (.onnx). With this, all trained or retrained models and parameter artifacts are ready to be exported and deployed to test or production environments.
\end{itemize}

\subsubsection{Decision making} 
A properly designed decision making strategy is key to making a system interact with the environment safely. Our strategy was to detect when air quality anomalies occur. The anomalies preceded a situation when a particular room developed uninhabitable conditions. Machine learning models performs regression and a separate layer then detects anomalies.

Evaluation of the strategy was done based on model and system performance. We decide in terms of accuracy of decisions and their usefulness to improve it. From Table \ref{tab:model training results}, we can observe the accuracy of decisions made by the models in terms of the RMSE score. When the detected RMSE value was above 10, a new model was trained on more recent data and deployed to ensure optimal decision making and functioning.

\subsubsection{Automated accountability}
Automated systems enable continuous operations of the system without human or other dependencies. Automation for machine learning based systems is driven by seamless monitoring, continuous integration and continuous delivery as following:

\paragraph{Continuous Integration (CI) and Continuous Delivery (CD)}

\begin{figure}[t]
    \centering
    \captionsetup{justification=centering}
    \includegraphics[width=9cm]{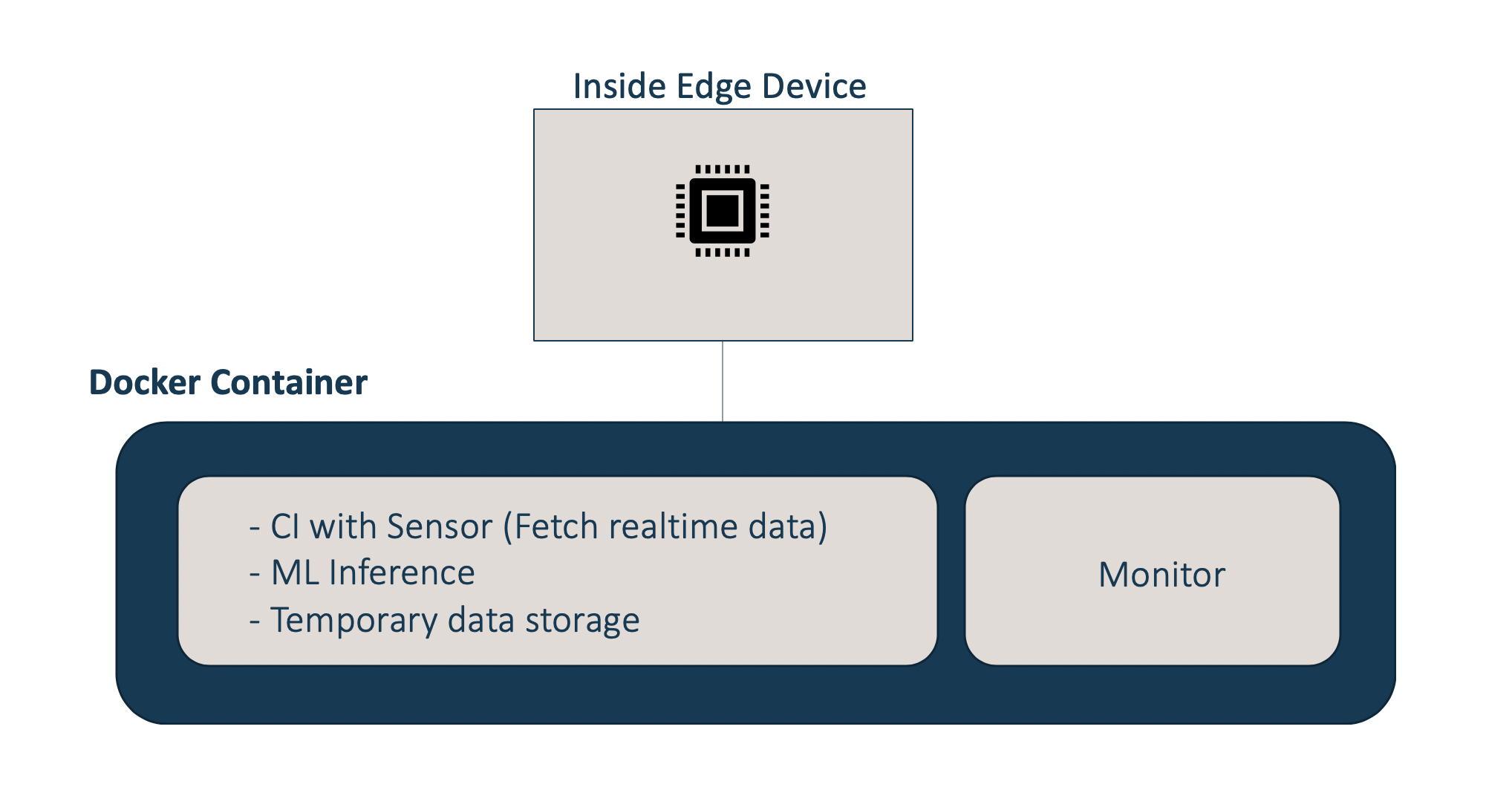}
    \caption{Docker container deployed in each edge device.}
    \label{Figure:docker-process}
\end{figure}

Our system is based on multiple edge devices by using continuous integration to ensure model and device freshness. In order to have a seamless continuous integration, two scripts or processes are running inside the docker container deployed in each edge device, as shown in figure \ref{Figure:docker-process}. These processes orchestrate data pipelines, machine learning, continuous integration, and deployment. The activities of re-training ML models, inference, and monitoring are automated as part of continuous delivery and deployment operations. The two processes are running inside a docker container on each edge device. This way of working is found to provide a reliable system while also being scalable. However, we must note that the implementation is still being revised and improved as this is a prototype. In table \ref{tab:ML-inference}, we show the run-time monitoring events that have been detected and handled, as explained in the processes below. 

\emph{Process 1}: This process enables and maintains sensor-to-edge continuous integration by fetching data in real-time. This is done by subscribing to a sensor topic using MQTT protocol. After new data is received from a sensor (which happens every 5 minutes), raw data is pre-processed by discarding or pruning unnecessary data, cleaning, and converting data into features. 

A machine learning model previously trained in the cloud is deployed to the edge device inside a docker container. The inference is then made to predict air quality 15 minutes into the future based on variables extracted from sensor data: air quality, ambient light, humidity, iaq accuracy static, pressure, and temperature. After getting a prediction for the real-time data, both sensor data and prediction are concatenated together and appended to a .csv file temporarily stored in the docker container.

\emph{Process 2}: This process is triggered for monitoring ML model performance at a set time every day (time trigger). When activated, the process evaluates the model drift by evaluating the RMSE for future air quality predictions vs. actual data. If RMSE is greater than or equal to 10, it means that model performance is poor. Hence the process evokes a call to look for and deploy an alternative model from the ML model repository on the cloud.

\begin{table}
 \caption{ML INFERENCE, CONTINUOUS DELIVERY AND RETRAINING RESULTS.}
 \centering
\begin{tabular}{ |p{0.5cm}||p{1.5cm}|p{2cm}|p{0.7cm}|p{0.7cm}|p{0.7cm}|  }
 \hline
 \multicolumn{6}{|c|}{Realtime machine learning inference at the edge} \\
 \hline
 S.no & Date of model change & Edge Device & Deployed Model& Model Drift (RMSE) & Model Retrain (RMSE) \\
 \hline
 1 &   15-03-2020  & Jetson nano 2  &ELM & 16.39 & 4.1\\
 2   & 16-03-2020    &Google TPU edge &   RFR& 14.23 & 6.3\\
 3 &   16-03-2020  & Raspberry pi 4   &MLR& 11.91 & 4.3\\
 4   & 17-03-2020    &Raspberry pi 4 &   ELM& 13.27 & 8.1\\
 5 &   22-03-2020  & Jetson nano 2  &SVR & 22.32 & 6.2\\
 6   & 24-03-2020    &Google TPU edge &   RFR& 17.11 & 4.4\\
 7 &   27-03-2020  & Raspberry pi 4 &MLR & 16.22 & 4.7\\
 8   & 29-03-2020    &Jetson nano 2 &   ELM& 30.28 & 8.2\\
 9 &   30-03-2020  & Google TPU edge   &SVR & 18.12 & 5.4\\
 10   & 05-04-2020   &Raspberry pi 4 &   MLR& 12.92 & 3.2\\
 11 &   10-04-2020  & Jetson nano 2   &SVR & 17.21 & 5.2\\
 12   & 11-04-2020    &Google TPU edge &   MLR& 13.42 & 4.7\\
 13 &   13-04-2020 & Jetson nano 2  &ELM & 27.29 & 5.3\\
 14 & 17-04-2020  &Google TPU edge &   RFR& 17.46 & 6.9\\
 15 & 19-04-2020  & Raspberry pi 4 &SVR & 16.32 & 5.1\\
 16   & 19-04-2020    & Google TPU edge &  MLR& 11.91 & 3.4\\
 17 &   21-04-2020  & Jetson nano 2   &ELM & 23.26 & 7.3\\
 18   & 22-04-2020    &Google TPU edge &   RFR& 16.92 & 7.2\\
 19 &   24-04-2020 & Raspberry pi 4 &SVR & 17.87 & 5.2\\
 20   & 25-04-2020    &Google TPU edge &   MLR& 13.92 & 5.2\\
 21 &   25-04-2020  & Jetson nano 2   &SVR & 19.21 & 7.9\\
 22   & 26-04-2020    &Raspberry pi 4 &   ELM& 23.57 & 6.4\\
 23 &   26-04-2020  &Google TPU edge &SVR & 18.21 & 5.5\\
 \hline
\end{tabular}
    \label{tab:ML-inference}
\end{table}

\paragraph{Record keeping}
All the models deployed and re-trained are end-to-end traceable and reproducible. Auditing and record maintenance enable traceability, validation, explainability (which model is used at a particular time index), reproducibility, and ability to show compliance to data protection regulation. 

\subsection{Reliability of fleet analytics}

Fleet analytics for the experiment's duration was based on data collected, without any interruptions, from each edge device used in the experiment. Each device's data provided an overview of device performance, based on telemetry data like accelerometer, gyroscope, humidity, magnetometer, pressure, and temperature. Edge device performance was stable overall during the experiment. All decisions were monitored statistically based on defined metrics and thresholds; this enabled the system's comprehensive supervision. The analytics process was comprehensively monitored as part of fleet analytics, including model training performance and inference performance in production.

\subsubsection{Analytics Process} The process of model training, deploying on edge devices, and monitoring the models are covered by Fleet analytics. All models trained and deployed are end to end traceable and auditable in real-time, as seen in the results of the model drift and re-train experiments in Table \ref{tab:ML-inference}. All models trained, deployed, and monitored for fitness were successfully observed without any failures or anomalies. The analytics process implemented for the experiments was based on the strategy devised to make the air quality monitoring system work efficiently with real-time supervision for the analytics process and infrastructure monitoring enabled by fleet analytics. 

\subsubsection{Supervision}
\begin{figure}[tbp]
\includegraphics[width=8cm]{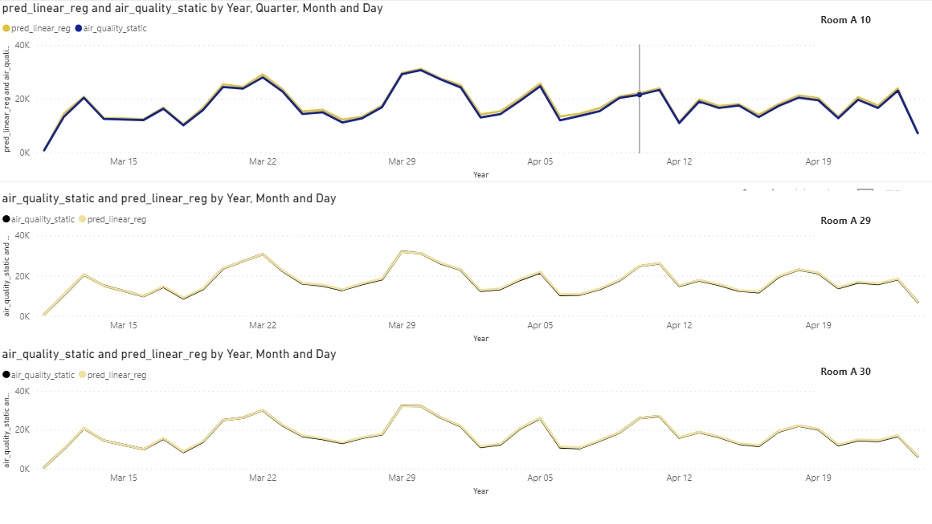}
\caption{Fleet Analytics - Analytics process}
\label{Analytics Process}
\end{figure}

System supervision is enabled statistical metrics defined to monitor the business problem. For our experiment, the business problem is forecasting future air quality, looking for signals, and alert using alarms to the building maintenance personnel. In case of future air quality forecasted above 100 aqi the system would alert the users (building maintenance personnel) to regulated air quality in the rooms. For machine learning models, a supervision threshold of 10 RMSE score was set. In case of RMSE crossing 10 RMSE at the end of the day then the model is replaced by another model and retrained on the latest data to improve the model for future use, this process of monitoring the models, deploying for replacing models, and retraining models are automated and enabled by continuous deployment. Fleet analytics (Analytics process) for models performance over time in three edge devices can be observed in Figure \ref{Analytics Process}.

\subsubsection{Device Analytics}

For each device, analytics provided an overview of device performance over some time with telemetry data like accelerometer, gyroscope, humidity, magnetometer, pressure, and temperature. Useful information to monitor edge devices health and longevity, all edge
devices' performance was stable overall throughout the experiment without any device failures.

% \begin{figure}[htbp]
% \centering%
% \includegraphics[width=8.7cm]{images/fleet analytics - devices.png}
% \caption{Fleet Analytics - Devices}
% \label{Telemetry data}
% \end{figure}

\section{Conclusion}\label{sec:conclusion}

Improving industrial processes using state-of-the-art analytics tools is a challenge despite the plethora of technological advances in IoT. This situation encourages the development of new frameworks with the capacity to bring stability and reliability. This paper presented a novel fleet analytics framework for handling edge IoT devices to improve the decision making process's fleet analytics. Our architecture also allows the user to optimize and scale the process with ease. We tested our framework by four different ML models on three different IoT devices to predict the air quality conditions in different rooms. The obtained results show that our approach is stable and reliable, and the retraining process and deployment was achieved without failure in all edge devices. In the future, we aim to consider scaling targets such as optimization of costs, operational clarity, and resource utilization to facilitate efficient edge-cloud operations at scale. We also plan to explore generalized metrics to evaluate the performance of the proposed framework.

\section*{Acknowledgments}
E.R. would like to thank TietoEVRY and the 5G-Force project funded by Business Finland. M.W. and L.E.L. thank the support of the Finnish Ministry of Education via the Master ICT for funding.

% Finland (Finland 2020)
%This paper is supported by TietoEVRY and 5G-Force project (5G-Force). 
%5G-Force is
%part of 5G Test Network Finland (5GTNF 2020). 5G-Force project is funded by Business
%Finland (Finland 2020). Thank you TietoEvry and 5G-Force project for the support.

%%%%%%%%%%%%%%%%%%%%%%%%%%%%%%%%%%%%%%%%%%%%%%%%%%%%%%%%%%%%%%%%%%%%%%%%%%%%%%%%%%%%%%%%%%%%%%%%%%%%

\bibliographystyle{IEEEtran}
\bibliography{main}

\end{document}